\documentclass[preprint,5p,times,twocolumn]{elsarticle}

\usepackage{amssymb}
\usepackage{lineno}
\usepackage{graphicx}
\usepackage[]{hyperref}
\usepackage{url}
\usepackage{array}
\usepackage{verbatim}
\usepackage{textcomp}
\usepackage{multirow}
\usepackage{tabularx}
\usepackage{textcomp}
\usepackage{verbatim}
\usepackage{subcaption}
\biboptions{sort&compress}
\usepackage{algorithm2e}
\usepackage{amsmath}
\usepackage{makecell}

\journal{Energy and Buildings}

\makeatletter
\renewcommand{\@algocf@capt@plain}{above}% formerly {bottom} Placing caption above

\def\@author#1{\g@addto@macro\elsauthors{\normalsize%
		\def\baselinestretch{1}%
		\upshape\authorsep#1\unskip\textsuperscript{%
			\ifx\@fnmark\@empty\else\unskip\sep\@fnmark\let\sep=,\fi
			\ifx\@corref\@empty\else\unskip\sep\@corref\let\sep=,\fi
		}%
		\def\authorsep{\unskip,\space}%
		\global\let\@fnmark\@empty
		\global\let\@corref\@empty  %% Added
		\global\let\sep\@empty}%
	\@eadauthor={#1}
}
\makeatother

\begin{document}

\begin{frontmatter}

%% Title, authors and addresses

%% use the tnoteref command within \title for footnotes;
%% use the tnotetext command for theassociated footnote;
%% use the fnref command within \author or \address for footnotes;
%% use the fntext command for theassociated footnote;
%% use the corref command within \author for corresponding author footnotes;
%% use the cortext command for theassociated footnote;
%% use the ead command for the email address,
%% and the form \ead[url] for the home page:
%% \title{Title\tnoteref{label1}}
%% \tnotetext[label1]{}

\title{Development and application of a machine learning supported methodology for measurement and verification (M\&V) 2.0\tnoteref{t1}}
\tnotetext[t1]{\textcopyright 2018. This manuscript version is made available under the CC-BY-NC-ND 4.0 licence \url{https://creativecommons.org/licenses/by-nc-nd/4.0/}}
% Decide whether to use based or supported in title

\author[ucc,marei]{Colm V. Gallagher\corref{cor1}}
\ead{c.v.gallagher@umail.ucc.ie}
\ead[url]{www.ucc.ie/en/ierg}
\cortext[cor1]{Corresponding Author}

\author[ucc,marei]{Kevin Leahy}
%\ead{kevin.leahy@umail.ucc.ie}

\author[ucc]{Peter O'Donovan}
%\ead{peter_odonovan@umail.ucc.ie}

\author[ucc,cit]{Ken Bruton}
%\ead{ken.bruton@cit.ie}

\author[ucc,marei]{Dominic T.J. O\textquoteright Sullivan}
%\ead{dominic.osullivan@ucc.ie}

\address[ucc]{Intelligent Efficiency Research Group, School of Engineering, University College Cork, Cork, Ireland}
\address[marei]{MaREI Centre, Environmental Research Institute, University College Cork, Cork, Ireland}
\address[cit]{Department of Mechanical, Biomedical and Manufacturing Engineering, Cork Institute of Technology, Cork, Ireland}

%% use optional labels to link authors explicitly to addresses:
%% \author[label1,label2]{}
%% \address[label1]{}
%% \address[label2]{}

\begin{abstract}
The foundations of all methodologies for the measurement and verification (M\&V) of energy savings are based on the same five key principles: accuracy, completeness, conservatism, consistency and transparency. The most widely accepted methodologies tend to generalise M\&V so as to ensure applicability across the spectrum of energy conservation measures (ECM's). These do not provide a rigid calculation procedure to follow. This paper aims to bridge the gap between high-level methodologies and the practical application of modelling algorithms, with a particular focus on the industrial buildings sector. This is achieved with the development of a novel, machine learning supported methodology for M\&V 2.0 which enables accurate and reliable quantification of savings.

A novel and computationally efficient feature selection algorithm and powerful machine learning regression algorithms are employed to maximise the effectiveness of available data. The baseline period energy consumption is modelled using artificial neural networks, support vector machines, k-nearest neighbours and multiple ordinary least squares regression. Improved knowledge discovery and an expanded boundary of analysis allow more complex energy systems be analysed, thus increasing the applicability of M\&V. A case study in a large biomedical manufacturing facility is used to demonstrate the methodology's ability to accurately quantify the savings under real-world conditions. The ECM was found to result in 604,527 kWh of energy savings with 57\% uncertainty at a confidence interval of 68\%. 20 baseline energy models are developed using an exhaustive approach with the optimal model being used to quantify savings. The range of savings estimated with each model are presented and the acceptability of uncertainty is reviewed. The case study demonstrates the ability of the methodology to perform M\&V to an acceptable standard in challenging circumstances. 
\end{abstract}


\end{frontmatter}
%\linenumbers

\section{Introduction}
\label{intro}
%SHORTCOMINGS IN CURRENT METHODOLOGIES
There are many widely recognised and well established methodologies for the measurement and verification (M\&V) of energy savings. These include the International Performance Measurement and Verification Protocol (IPMVP) \citep{EfficiencyValuationOrganization2016}, the American Society of Heating, Refrigerating and Air-Conditioning Engineers' (ASHRAE) Guideline 14 \citep{ASHRAE2014} and ISO 50015-2014 \citep{ISO50015}. These methodologies are intertwined with one another and provide guidance on applying universal approaches to the wide spectrum of energy saving projects. Despite this, the lack of a rigid calculation process has been highlighted as a significant shortcoming of these methodologies \citep{Ginestet2010a}. This is less of an issue in residential and commercial applications as the nature of the energy systems in place are more simplistic. In contrast to this, industrial buildings contain complex energy systems with many variables impacting on energy consumption. ASHRAE Guideline 14 states that its procedures do not include major industrial loads. The lack of a prescribed, analytical process that can be applied has implications on the accuracy and reliability of energy savings.

%IMPACT ON EU POLICY
In 2015, industry accounted for 25.3\% of total final energy consumption in the European Union (EU) \citep{Eurostat2016} and 20.9\% in Ireland in 2016 \citep{SEAI2017}. The European Parliament have issued the Energy Efficiency Directive in an attempt to maximise the efficiency with which energy is consumed in industry \citep{EuropeanParliament2012}. Under the terms of the Directive, member states are obligated to achieve 20\% energy efficiency savings by 2020. The success of energy conservation measures (ECMs) implemented to achieve this target can only be measured using M\&V. Thus, accurate M\&V is a necessity for ECMs to be confidently relied upon when assessing progress towards EU targets. The focus on energy efficiency is set to continue post-2020 with proposals to update the Directive that include a new 30\% energy efficiency target for 2030 \cite{Commission2016}. M\&V has a critical role to play in the delivery of this energy efficiency policy. 

%CUMULATIVE IMPACT OF ECMS & PAPER 1 FINDINGS
The major focal point of legislation has been the implementation of ECMs to minimise consumption in the industrial sector. For this to be a success, the M\&V performed on each individual ECM must be of sufficient accuracy so that the savings estimated can be relied upon. The cumulative impact of these ECMs will be evidence of the success of the Directive. There is a significant danger that over estimation of savings on an individual project level could hinder attempts to limit climate change. This has created a need for a methodology that is capable of overcoming the barriers that impede accurate M\&V in industrial facilities. These challenges include cost, resources and the time required to perform M\&V. Earlier research aimed at addressing these issues identified the potential of machine learning as a suitable tool to achieve this \cite{Gallagher2018}; hence, this paper proposes a machine learning supported methodology to populate this knowledge gap. A clearly defined, prescriptive process focused on harnessing the power of energy data in an efficient manner is presented. 

%CONCEPT OF M&V AND UNCERTAINTY
There are three periods of interest in any M\&V project: the baseline, implementation and reporting periods. Although they always occur sequentially, the length of each period will vary depending on individual project parameters. The baseline period occurs prior to the implementation of an ECM, with the reporting period taking place following the implementation period. A crucial step in M\&V is the estimation of the adjusted baseline in the reporting period. This is found by normalising the reporting period energy consumption to baseline period conditions. Typically, engineering or statistical methods are applied to construct a baseline model capable of performing this normalisation. Consequently, M\&V is not an exact science and maintaining accuracy throughout the process is critical to its success. The three principle sources of uncertainty are sampling, measuring and modelling. The IPMVP defines a methodological approach that can be applied to quantify uncertainty in a project. In this, the minimum acceptable level of uncertainty is defined as the point at which savings are larger than twice the standard error of the baseline value \citep{EfficiencyValuationOrganization2014}. Lee et al. identified uncertainty of baseline measurements as a key risk to energy service companies in energy performance contracting \citep{Lee2015}.

%\textbf{NEED TO INCLUDE PLACE FOR M\&V 2.0 IN INDUSTRY 4.0 HERE}
In recent years, M\&V 2.0 represents an area of significant interest and it is being used to further develop the commonly used practices. M\&V 2.0 differs from traditional M\&V as it uses large data sets and automated advanced analytics to streamline and scale the process \citep{Grandersona2017}. The automated analytics can provide ongoing savings estimates in close to real-time. This enables M\&V to progress from a static, retrospective process to a more dynamic state in which savings can be maximised. The added complexity of modelling has been driven by the increased availability of granular energy data from advanced metering infrastructure (AMI) systems. The use of this data coupled with automated processing has been identified as the most opportune manner with which to progress M\&V \citep{Franconi2017}. The increased accuracy, certainty and standardisation of savings calculations offered by M\&V 2.0 is hugely beneficial. To enable this, there is a need to establish guidelines and best practices in order to fully realise the potential of these advancements. The formal methodology presented in this paper seeks to become this resource for the industrial sector. 

%The remainder of this paper is organised as follows: Section \ref{goals} outlines the authors' research objectives, Section \ref{related} presents an overview of related work published in this field, Section \ref{methodology} defines the proposed methodology, Section \ref{casestudy} presents the applications, Section \ref{results} contains the results and Section \ref{concs} includes the conclusions drawn from this research. 

\section{Research questions}
\label{goals}
To date, artificial intelligence (AI) has been proven to be advantageous in building energy load prediction \citep{Wang2017}, with machine learning being a sub-field of AI. The primary objective of this research is the development of a replicable, robust and detailed methodology to enable the use of machine learning for the purposes of M\&V in industrial facilities. The following research questions were used to lead the analysis:

\begin{enumerate}
	\item Can a definitive methodology be developed to provide explicit guidance on the application of machine learning in M\&V? 
	\item Is it possible for such a methodology to be robust enough to harness the power of available data across the spectrum of different M\&V projects?
	\item Can machine learning algorithms be employed on large data sets without increasing the resources required for M\&V?
	\item An extended boundary of analysis is proposed to increase the baseline energy model accuracy in circumstances with limited system specific metering infrastructure. Can M\&V be completed with acceptable accuracy using this novel boundary of analysis?
	%\item What is the trade-off between accuracy and the resources required to achieve the accuracy?
\end{enumerate}

\section{Related work}
\label{related}
%Each subsection will lead the review on a different topic. The areas to be covered are energy savings verification in industrial facilities and the novel approaches employed to date, machine learning in M\&V, and the advantages of ML with respect to temporal granularity and missing data. 

\subsection{Methodologies}
\label{methods}
%PARAGRAPH 1: The Big 3
As introduced in Section \ref{intro}, the IPMVP is the most prevalent M\&V methodology employed worldwide. Four distinct approaches are defined to cover a wide range of projects. Options A and B isolate the retrofit with a project boundary that encompasses the affected equipment. Option C is a whole-building approach and applicable in cases where the savings are greater than 10\% of the total site energy consumption. Option D consists of a calibrated simulation of the energy systems. This approach is beneficial in circumstances with no baseline data. ASHRAE Guideline 14 and ISO50015 present methodologies that are based on the same core concepts as the IPMVP. These generalised methodologies have the distinct benefit of being robust. The most significant drawback of this approach is the widely publicised lack of guidance on the calculation process. This issue is amplified when performing M\&V in industrial facilities, where the quantity of factors impacting on energy performance complicates the modelling process. 

%PARAGRAPH 2: Novel Methodologies
There have been a variety of alternative approaches proposed to expand the knowledge base in the industrial sector and attempt to overcome these aforementioned problems. Therkelsen et al. analysed data from five industrial buildings to compare absolute, intensity and regression approaches to M\&V and found regression based approaches were the most effective  in translating energy savings values into contextualised energy performance improvement values \citep{Therkelsen2016}. Kelly Kissock and Eger proposed a general methodology that takes weather and production into account for measuring plant-wide energy savings \citep{KellyKissock2008}. This approach also has the ability to disaggregate savings into components which provides additional resolution. This offers a novel, alternative approach to the traditional methodologies; however, the method was found to be limited by the information in the data set, which can be sparse as whole-facility billing data is used. Rossi and Vel{\'{a}}zquez presented a prescriptive methodology for performing M\&V on combined heat and power (CHP) plants in industrial facilities \citep{Rossi2015}. Burman et al. developed an M\&V plan to compare the design and actual energy performance of a building. This approach was proposed with a view to identifying the shortcomings in the construction and building procurement processes \citep{Burman2014b}. 

%Uncertainty
The traditional methodologies also provide guidance on quantifying uncertainty associated with estimated energy savings \citep{EfficiencyValuationOrganization2014}. Granderson et al. developed an alternative statistical methodology that can be used to evaluate the accuracy of building energy models. It should be noted that this approach uses data from buildings in which no ECM was implemented to quantify uncertainty and offers a useful alternative for comparing the performance of modelling algorithms. The five models evaluated were all found to perform poorly when energy use varied in ways that were not predictable from the outdoor air temperature or the time of week \citep{Granderson2014}. This further exposes the need to develop more intricate approaches to improve the applicability to the industrial sector. 

%PARAGRAPH 3: Automated Approaches
In recent years, M\&V 2.0 tools have been developed in an attempt to automate the process. Ke et al. created a cloud-based platform for estimating energy savings for any ECM with a case study in a commercial building \citep{Ke2017}. Granderson et al. applied an automated whole-building M\&V tool to historic data sets from energy efficiency programs and performed a comparison with traditional methods \citep{Grandersona2017}. Ke et al. also applied eQUEST software to calibrate energy simulation results using IPMVP Option D \citep{Ke2013a}. Advancing the algorithms used for energy modelling in industrial buildings is the first step in evolving to M\&V 2.0 practices in the industrial sector. 

\subsection{Baseline energy modelling}
\label{algorithms}
The critical step in any M\&V methodology is the development of an energy model prior to the implementation of the ECM. This is known as the baseline energy model. It has been highlighted in Section \ref{methods} that the approaches taken to model this baseline energy consumption need to evolve in order to fully capture the behaviour of systems in industrial buildings. The use of more complex algorithms, beyond that of the typical ordinary least squares linear regression techniques commonly applied, enables practitioners to accurately model the complex energy consumption behaviour in these scenarios; hence improving the certainty with which savings are quantified. 

%PARAGRAPH 1: Baseline Energy Modelling in the Context of M\&V (Focus on industry, but highlight the shortcomings of other ones)
Heo and Zavala utilised Gaussian process (GP) modelling to determine energy savings and uncertainty levels in commercial office buildings. The GP models developed were capable of capturing the complex non-linear and multi-variable interactions, as well as multi-resolution trends of energy behaviour \citep{Heo2012}. Sever et al. proposed inverse simulation as a less time-intensive method of estimating energy savings in the industrial sector. Savings were determined using a multi-variate three-parameter change-point regression model driven with typical weather data \citep{Sever2011}. Granderson et al. reviewed 10 different approaches to baseline energy modelling for whole-building M\&V methods in the commercial sector; the findings of which can be used to build confidence in model robustness. The models tested utilised a variety of different algorithms including principle component analysis, random forests, mean-week and time approaches, advanced regression and nearest neighbours \citep{Granderson2016}. Dong et al. employed support vector machines (SVM's) to forecast the energy consumption of four commercial buildings using outdoor air temperature, relative humidity and global solar irradiation as predictor variables \citep{Dong2005a}. Rossi et al. applied artificial neural networks (ANN) to accurately model the baseline energy consumption of CHP plants \citep{Rossi2014a}. 

%PARAGRAPH 2: Introduce approaches taken to model energy consumption in industrial facilities without the context of M\&V.
%FIRST PART: All reviews
Energy modelling has a whole host of applications that lie outside of the scope of M\&V. Therefore, it is prudent to review the approaches taken elsewhere in an effort to advance the research field of interest. As energy modelling in buildings is quite a mature field, there have been a number of extensive reviews carried out. Zhao and Magoul{\'{e}}s conducted a comprehensive review of simplified engineering, statistical and AI methods for the modelling and prediction of energy consumption in buildings \citep{Zhao2012}. Yildiz et al. conducted a review of regression and machine learning models for electricity load forecasting in commercial buildings \citep{Yildiz2017a}. Harish and Kumar carried out a comprehensive review of modelling and simulation techniques for building energy systems, which included space loads, heating ventilation and air conditioning systems and lighting \citep{Harish2016b}. Foucquier et al. reviewed machine learning, thermal and hybrid approaches used to model energy consumption, heating/cooling demand and indoor temperature in buildings. It was concluded that machine learning approaches can be the easiest to deploy, with the hybrid machine learning and thermal modelling approaches offering significant promise in the future \citep{Foucquier2013}. Tardioli et al. reviewed the recent application of data-driven models at an urban level and proved that they are useful in reducing the time taken to create an energy consumption model while maintaining an adequate level of accuracy \citep{Tardioli2015}. Finally, Ahmad et al. reviewed ANN and SVM's with the aim of identifying better forecasting of building's electricity consumption. This study concluded that artificial intelligence based approaches (i.e. machine learning) are advantageous in capturing the behaviour of complex energy systems influenced by many parameters \citep{Ahmad2014a}. 

%PARAGRAPH 3: A few specific examples of applications. 
There are several examples of machine learning being used to successfully predict energy consumption and related factors in buildings. Wei at al. carried out a multi-objective optimization of a heating, ventilation and air conditioning (HVAC) system performance using a data-driven approach \citep{Wei2015, Tang2014}. The multi-layer perceptron ensemble approach was used to build accurate models that considered both energy consumption and local environmental conditions such as air quality and CO\textsubscript{2} levels. Esen et al. have shown the successful use of ANN, SVM and adaptive neuro-fuzzy inference systems to model and predict the performance of ground-source heat pump systems with minimal data \citep{Esen2008d,Esen2008,Esen2008b,Esen2008c}. Similarly, SVM, ANN and wavelet neural network approaches have been applied to model solar air heaters, an otherwise difficult task when conventional methods are used \citep{Esen2009,Esen2009a}. On a less granular scale, Fumo et al. propose a simplified method for estimating hourly energy consumption of a building by applying a series of predetermined coefficients to the monthly energy consumption data from utility bills \citep{Fumo2010}. Carpenter et al. used billing data to develop GP and three-parameter cooling change point models \citep{Carpenter2016}. 

AI modelling algorithms, including machine learning approaches, have been proven to be beneficial for forecasting energy consumption in buildings. Within the scope of M\&V, the majority of research published to date focuses on residential and commercial buildings. This has led to a knowledge gap surrounding the application of these complex algorithms in industrial facilities for the purposes of energy savings verification. The methodology proposed and applied in this paper satisfies this need, which enables M\&V in industrial facilities to progress to M\&V 2.0. 

\section{Methodology}
\label{methodology}
AMI is common place in modern industrial buildings. These systems collect large quantities of granular energy data, which can be used to discover knowledge on system behaviour. This data is widely available, however, it is rarely utilised to its full potential. Common issues that hinder the use of this data include the lack of a central data repository, inefficient preprocessing techniques and insufficient subject matter knowledge. This data can significantly improve the accuracy with which M\&V can be performed, although the need for skilled professionals to perform tasks such as data cleaning and baseline energy modelling impedes the process. 

The methodology presented in this paper is capable of overcoming the issues impeding the effective use of available data. A novel alternative to the traditional M\&V protocol is offered. In contrast to IPMVP and ASHRAE Guideline 14, a prescriptive data handling and modelling procedure is detailed to ensure that maximum accuracy is achieved. In addition to this, the prescription of this methodology contributes to simplified decision making and reduces the need for subject matter expertise. 

\begin{figure*}[]
	\centering
	\includegraphics[height=0.98\textheight,keepaspectratio]{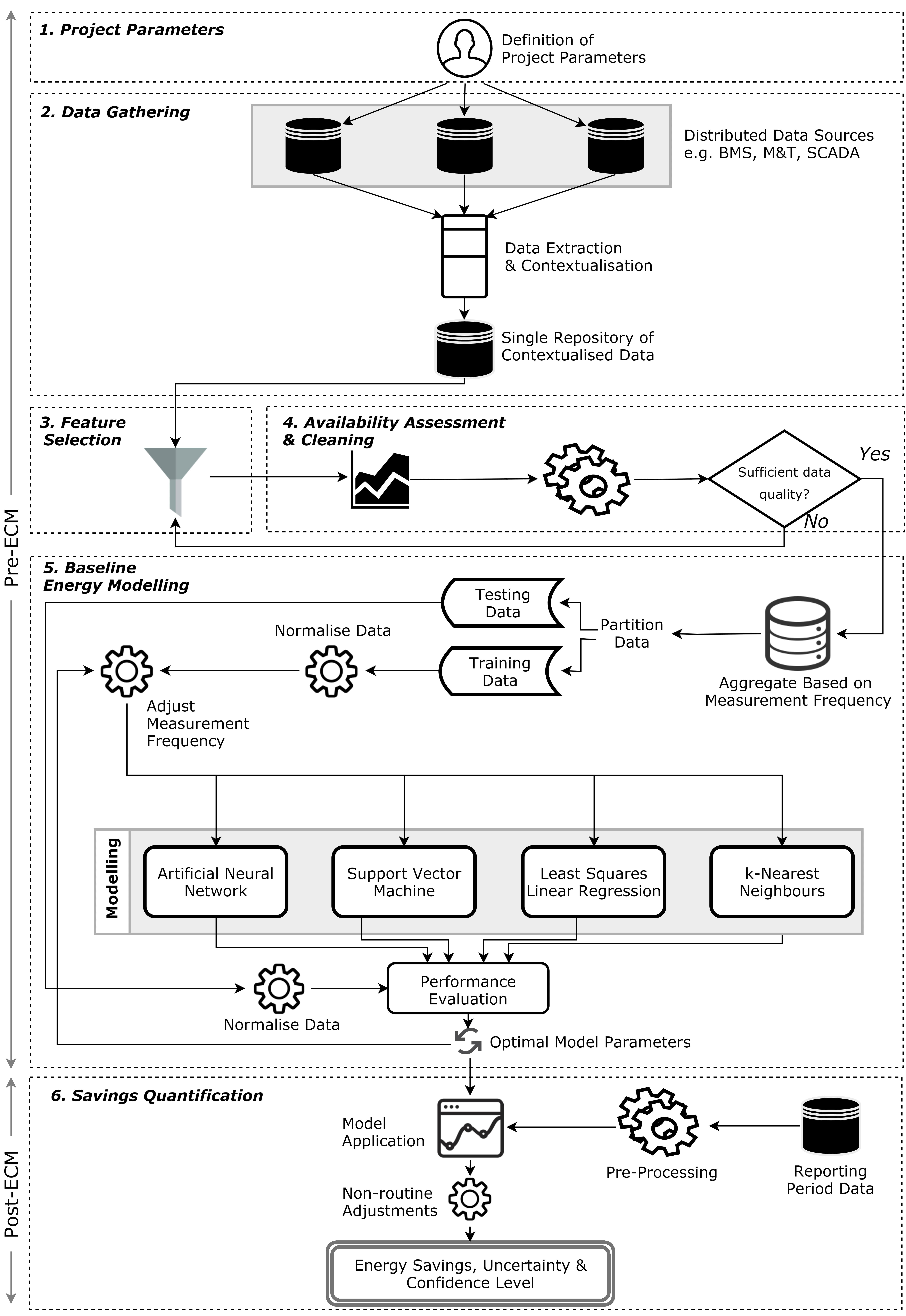}
	\caption{Process flow diagram of the proposed methodology.}
	\label{pfd}
\end{figure*}

It is important to note that the use of additional modelling techniques, not included in this methodology, is at the discretion of the M\&V practitioner. Additionally, the data handling framework is not completely prescriptive so as to maintain applicability and remain largely technology agnostic. More detailed data pipelines have been developed for data-driven analytics applications in large-scale industrial facilities and these can be integrated into the proposed methodology by the user \citep{ODonovan2015}. Figure \ref{pfd} illustrates the process flow diagram of the methodology.

\subsection{Step 1 - Definition of project parameters}
\label{define}
It is critical that the scope and boundaries of the project are defined prior to any commencement of work. This ensures that the M\&V of the resultant energy savings can be completed in an accurate, complete, conservative, consistent and transparent manner. To do this, the following items need to be documented;
\begin{itemize}
	\itemsep0em 
	\item ECM's to be carried out
	\item Project boundary
	\item Relevance to total energy consumption on entire site
	\item Project time-line: Expected baseline, implementation \& reporting periods
	\item Relevant personnel
	\item Data sources
	\item Static factors
\end{itemize}

\subsection{Step 2 - Data gathering}
\label{gathering}
\subsubsection{Extraction}
The characteristics of each relevant data source identified must be detailed. This should include the type of data, measurement frequency, storage methods and access protocol. The objective of this stage in the process is to outline a means of accessing data from each distributed data source to enable data extraction. Information on how to manage and access this data beyond the baseline period must also be included. As per ISO 50015, this includes, but is not limited to, storage, backup, maintenance and security of the data. The information collated at this stage should be replicable so that the same process can be followed during the reporting period. 

\subsubsection{Contextualisation}
\begin{figure}[h!]
	\centering
	\includegraphics[width=0.5\textwidth]{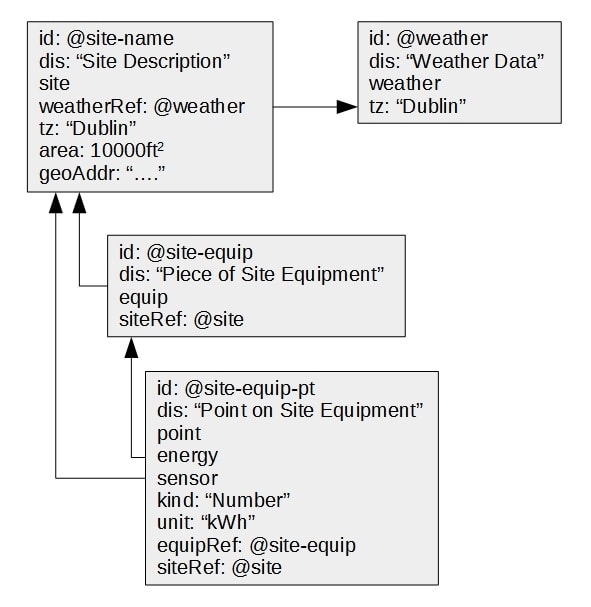}
	\caption{Basic three level hierarchy of the Haystack naming convention.}
	\label{haystack}
\end{figure}

Contextualisation of the relevant data is critical in gaining meaningful insights into the systems being analysed. Poor semantic modelling is common for energy data across the industrial sector. This has led to the need for a standardised methodology for describing data. One such solution to this problem is Project Haystack \citep{haystack}. The goal of the Haystack naming convention is to make it easier and more cost effective to analyse, visualise and derive value from operational data. The object oriented class hierarchy, illustrated in Figure \ref{haystack}, is based on three entities: the site, pieces of equipment and points. The naming convention uses a tag model to describe data within the context of the facility. A full reference guide for applying the naming convention is available online \citep{haystack}.

The Haystack naming convention is endorsed in this methodology as it has many uses far beyond the M\&V application. If deployed across systems on a site, communication and data accessibility is naturally improved, and hence analytics, is made more straightforward. In the context of this methodology, other naming conventions can be applied if desired. This step is important to ensure a contextualised, more useful data set is produced. 

\subsection{Step 3 - Feature selection}
\label{featsel}
The contextualised data set can often be very large as hundreds of variables are stored. It is important that only those variables that offer importance in model construction are brought forward for analysis. Therefore, feature selection is used to select a subset of relevant variables for use in model construction. 

To clarify, the term "variable" is used to refer to the raw input data and the term "feature" is used for those variables output from the feature selection process. These will be the features used at a later stage to construct the baseline energy consumption models. Feature selection has many benefits in this application, including reducing the measurement and storage requirements in the reporting period, minimising model training time and avoiding high dimensionality for improved prediction performance \cite{Guyon2003a}. 

Feature selection algorithms can be categorised as either filter or wrapper approaches. Filter-based solutions are generally more applicable to high dimensional data as they rely on the characteristics of the training data. In contrast to this, the wrapper approaches employ a learning algorithm and determines its performance to select the variable subset. This generally results in longer processing times and requires more computing resources. A combination of simple filter and wrapper approaches are employed in this methodology to minimise processing speed, while selecting the optimal feature subset. A Spearman rank correlation filter method is included in a wrapper that seeks to maximise the adjusted coefficient of determination, $R^2$, of the data set. This is found by constructing a multiple ordinary least squares (OLS) regression model. This approach measures the strength and direction of monotonic association between two variables using Algorithm \ref{spearman}. The Spearman correlation coefficient is defined as the Pearson correlation coefficient between two ranked variables. This is then used to iteratively select variables to add to the feature subset. The addition of each feature is evaluated with respect to the adjusted $R^2$ value produced using the previous feature subset. Algorithm \ref{fsalgor} details this iterative process which determines the optimal subset of features. 

\begin{algorithm}
	\caption{Calculation of Spearman rank correlation.}
	\label{spearman}
	\KwIn{Input data set expressed as an $m$ x $n$ matrix.}
	$x[\ ,n] = $ dependent variable \\
	\For{$j = 1, \dots, n-1$}{
	    $r_j = $ rank x[ ,j] \\
	    $r_n = $ rank x[ ,n] \\
		
		\eIf{All m ranks are distinct integers}{
	    \For{$i = 1, \dots, m$}{
	    	Compute \\
	    		$d_i = r_j[i] - r_n[m]$
	    }
    
    	Compute \\
	    $\rho = 1 - \frac{6\Sigma d_i^2}{m(m^2-1)}$
    	}{
    	
    	\For{$i = 1, \dots, m$}{
    		Compute 
    		$\rho = \frac{cov(r_j, r_n)}{\sigma_{r_j} \sigma_{r_n}}$, where
    		
    		{$cov(r_j, r_n)$ is the covariance of the rank variables}\\
    		{$\sigma_{r_j} \& \sigma_{r_n}$ are the standard deviations of the rank variables}\\
    	}
    }
}
	\KwOut{Spearman correlation coefficients, $\rho$, for each input variable w.r.t. the dependent variable.}

\end{algorithm}

%A fast-correlation based filter solution for feature selection \cite{Yu2003}. Relevance and redundancy are critical.

\begin{algorithm}
	\caption{Spearman rank-based feature selection to optimise adjusted multivariate coefficient of determination.}
	\label{fsalgor}
	\KwIn{$m$ x $n$ matrix containing all input data.}
	$x[\ ,n] = $ dependent variable \\
	Apply algorithm \ref{spearman} to calculate variable $ranks$ \\
	Order columns in input matrix by decreasing $\rho$ \\
	
	$ i = 1$
	
	$\rho_i =$ Spearman correlation coefficient between variable i and $x[, n]$
	
	$subset_i = x[\ , cols(1,\dots,i,n)]$, i.e. variable with highest $\rho$ and dependent variable
	
	Train OLS regression model for $subset_i$ \& find $r^2_{adj_i}$
	
	\While{$i$ != no. of variables}{
		$subset_{i+1} = x[\ , cols(1,\dots,i+1, n)]$ \\
		Train OLS regression model for $subset_{i+1}$ \& find $r^2_{adj_{i+1}}$ \\
		\If{$r^2_{adj_{i+1}} - r^2_{adj_i} > 0.01$}{
		$subset_{i} = subset_{i+1}$ \\
		$ r^2_{adj_i} = r^2_{adj_{i+1}}$ \\
		$i = i + 1$
		}
		\Else{
		Remove variable $i + 1$ from the dataset
		}
	\Return{$subset_{i}$}
	}
	
	\KwOut{Dataset with features selected.}
\end{algorithm}

To avoid multicollinearity, it is critical that the features identified are not just independent of the dependent variable, but also independent of each other. Multicollinearity can cause coefficient estimates in multiple regression models to change erratically in response to small changes in the model or data. The variance inflation factor (VIF) is used to test for multicollinearity between features and hence, avoid selecting redundant features. Similar to the coefficient of determination, there is no single value for the VIF that indicates multicollinearity. A commonly used rule of thumb is a VIF value greater than 10 indicates multicollinearity. For weaker models, values above 2.5 may cause concern. In keeping with the M\&V principle of conservatism, any feature found to have a VIF greater than 5 should be removed from the feature set. 

\subsection{Step 4 - Availability assessment and cleaning}
\label{avail}
A data availability assessment consists of an initial, high-level statistical analysis of the proposed model features. The results of this allow the practitioner to make an informed decision, based on data quality and integrity, as to which features are suitable for analysis. The simple summary statistical measures detailed in Table \ref{sumstats} are to be used to enable evidence based decision making.

\begin{table}[h!]
	\small
	\def\arraystretch{1.5}
	\begin{tabular}{p{0.12\textwidth}p{0.3\textwidth}}
		\hline
		Measure & Description \\
		\hline
		Mean & The average value in a set of numbers \\
		Median & The value lying at the midpoint of a frequency distribution of values. \\
		No. of Unique Values & The number of unique values in the set of measures for a variable. \\
		No. of Missing Values & The quantity of values missing the dataset. This is usually assessed with respect to a measurement frequency being used in the analysis. \\
		Quartiles & The three points that divide the data set into four equal groups, each group comprising a quarter of the data, where the data is ordered sequentially. \\
		Minimum & The lowest value in the set. \\
		Maximum & The highest value in the set. \\
		\hline
	\end{tabular}
	\caption{Statistical measures to be employed in the data availability assessment.}
	\label{sumstats}
\end{table}

Features with large numbers of outliers, periods of missing data or unreliable measurements should be omitted from the feature subset. As a rule of thumb, features with more than 5\% of poor quality data should be omitted entirely from the subset. Any features that fall short of this 5\% omission threshold can generally be cleaned using the process detailed in Section \ref{avail}. In addition to the summary statistics, visualisation techniques can also be used to gain an understanding of the data at hand. Box plots, time series plots and histograms are useful in graphically representing the data. This process ensures that data quality and integrity is maintained. Section \ref{casestudy} presents an implementation of this assessment.

Data cleaning is the process of detecting and removing inaccurate entries in a data set. Maintaining quality in the baseline period data is critical to ensuring the system under analysis is accurately modelled. Under the IPMVP, baseline data should not be replaced by modelled data, except when using Option D \citep{EfficiencyValuationOrganization2012}. Therefore, the scope of data cleaning in this application is limited to simply identifying unclean data and subsequently removing it. No backfilling of data is to take place. The only exception to this is if data is missing for a consistent period of time in the baseline period, then comparable data for the same time period in a different year can be employed.

The results of the data availability assessment carried out in Section \ref{avail} are to be used to guide the data cleaning process. Using the summary statistics, box plots, time series plots and histograms output from the assessment, variables with irregularities can be identified. Some of these irregularities cannot be rectified with data cleaning; these features must be omitted from the analysis. If features must be omitted, the feature selection algorithm should be reapplied with these features removed from the data set. This can allow for new features to be included in the data set. 

\subsection{Step 5 - Baseline energy modelling}
\label{modelling}
The development of the optimal model of the baseline energy consumption is critical to ensuring that the uncertainty associated with the final energy savings is minimised. This model is referred to as the baseline energy model. An exhaustive process is used to ensure the model developed is tailored to the characteristics of each specific project. 

\subsubsection{Aggregate based on measurement frequency}
\label{freq}
Firstly, the data is aggregated based on the measurement frequency. This is necessary as each subsequent step is frequency specific. The objective of this step is to generate multiple data sets to enable an array of models be developed. The number of data sets that can be created is dependent on the frequency with which the data is measured. The wide availability of AMI in modern industrial buildings generally results in data being recorded in 15-minute intervals. For this case, the data is then aggregated using the mean values for hourly and daily measurement frequencies. This results in 3 data sets being available for model development purposes. It is not advised that less granular data than that with a weekly measurement frequency be used as these can result in insufficient quantities of testing data leading to unreliable results. 

\subsubsection{Partitioning of data}
\label{partition}
The data gathered at this stage of the process is for the baseline period only as the ECM has not yet been implemented. The data sets output from the previous aggregation stage is partitioned into training and testing data sets. This enables the models to be constructed using the training data and tested on an unseen set. A shuffled split is performed with 80\% of data used for training and 20\% used for testing. This is in contrast to the guidance given by the IPMVP which uses 100\% of the baseline data to construct the model and subsequently calculates the performance metrics by applying the model to the same data set \cite{EfficiencyValuationOrganization2014}. This approach may be prone to over-fitting the model to the training data, which can result in random error or noise being incorporated, resulting in unreliable performance evaluation. The partitioning of the baseline period data and testing on a data set not used in model construction may prove a more accurate approach that results in a reliable and independent evaluation of performance. 

The training data is brought forward to the next stage in the methodology, while the testing data set is not used again until performance evaluation is required in Section \ref{eval}.

\subsubsection{Feature scaling}
\label{scaling}
Feature scaling is used to standardise the range of features in the data set with a view to improving model performance. It also improves the processing time of certain algorithms including ANNs. This process is also known as standardisation or Z-score normalisation and results in each feature having the properties of a standard normal distribution (i.e. standard deviation of 1 and mean of 0). Each of the training data sets input into this stage of the process are standardised and the scaling parameters of each feature in these data sets are stored for application at a later stage. 

%Ref/Useful: \url{http://sebastianraschka.com/Articles/2014_about_feature_scaling.html}

\subsubsection{Model training}
\label{modtraining}
Baseline energy models are trained using the data sets for each measurement frequency. This is an exhaustive process which seeks to identify the most appropriate model hyper-parameters for each algorithm and measurement frequency. Section \ref{algorithms} reviews the success of various machine learning algorithms in the field of M\&V to date. The ability of machine learning algorithms to model energy systems in industrial buildings outside the context of M\&V is also presented. The findings of this review led to the following algorithms being deemed the most appropriate for application:
\begin{enumerate}
	\item Ordinary least squares regression
	\item k-Nearest neighbours regression 
	\item Artificial neural networks
	\item Support vector machine regression
\end{enumerate}

Each of the algorithms requires the values of certain parameters be set prior to the commencement of the learning process. These are known as hyper-parameters. A grid-search approach using 10-fold cross validation is employed to find the optimal values of each hyper-parameter. The recommended grid-search values for each hyper-parameter are detailed in Table \ref{hyper}. As this is not an exhaustive list of grid-search values, practitioners may choose to alter the specifications of each. The values provided are recommended based on previous research on this topic \cite{Gallagher2018}. The training of models is an optimisation solution which uses an iterative approach to arrive at the final values of the hyper-parameters. 

\begin{table*}[h!]
	\small
	\def\arraystretch{1.5}
	\begin{tabular}{>{\raggedright}p{0.15\textwidth}p{0.45\textwidth}>{\raggedright}p{0.15\textwidth}>{\raggedright\arraybackslash}p{0.18\textwidth}}
		\hline
		Algorithm & Description & Hyper-parameters & Grid Search \\
		\hline
		Bi-variable Linear Regression & An ordinary least squares approach assumed to be representative of typical M\&V practice. & Intercept & True/False \\ 

		Multi-variable Linear Regression & A more detailed ordinary least squares model constructed using 9 additional features from the available data set. & Intercept & True/False\\

		k-Nearest Neighbours & Non-parametric model where the input consists of the k closest training examples in the feature space. The output is the average of the values of its k-nearest neighbours. & 
		Maximum no. of neighbours\break  Distance\break Kernel	 & $k_{max} = $ 1:10\break\break $d = $ 1:5\break $kernel = $ triangular\\

		Artificial Neural Networks & Non-linear statistical model. It is a two-stage regression model typically represented by a network diagram. A single hidden layer feed-forward neural network was developed in each instance. & No. of hidden units\break Maximum no. of iterations\break Threshold\break Weight decay& $size = $ 1:10\break $it_{max} = $ 1,000\break\break $t = $ 0.01\break $d =$ (0.001,0.01,0.1,0.5) \\

		Support Vector Machines & Non-parametric technique reliant on kernel functions. Examples are represented as points in space with a clear gap separating mapping categories.& Kernel \break Cost & $kernel = $ linear\break $c = $ (0.25,0.5,1)\\
		\hline
	\end{tabular}
	\caption{Description of algorithms and associated hyper-parameters.}
	\label{hyper}
\end{table*}

A model constructed using each algorithm and measurement frequency is output from this stage. For a case assessing 3 measurement frequencies, there are 12 models developed.

\subsubsection{Performance evaluation}
\label{eval}
To identify the most suitable model, the testing data is used to evaluate the performance of each baseline energy model. The testing data sets defined in Section \ref{partition} are standardised using the scaling factors employed on the training data sets. These are specific to each modelling frequency. Each model developed is then applied to the appropriate standardised data set to produce a prediction of energy consumption. 

The coefficient of variation of root mean square error (CV(RMSE)) and normalised mean bias error (NMBE) are employed to quantify the prediction performance of each model. CV(RMSE) is a commonly employed performance metric and is used in both IPMVP and ASHRAE Guideline 14. It is a measure of the variability between actual and predicted values. The CV(RMSE) is employed as it gives context to the size of the error relative to the quantity being modelled. It is also important to note that the RMSE is known as the standard error (SE) in the IPMVP. 

NMBE is an indication of overall bias in a regression model. It quantifies the tendancy of a model to over or underestimated across a series of values. In contrast to the CV(RMSE), the NMBE is independent of time and hence, it can result in overall positive bias cancelling out negative bias. The use of both metrics in conjunction with each other allows for a true insight into model performance. 

Equations \ref{eq:cv} and \ref{eq:nmbe} are used calculate each metric, where $y_i$ is the actual value, $\hat{y_i}$ is the predicted value, $\bar{y}$ is the average of the actual value, and $n$ is the total number of predictions in the period of analysis.

\begin{equation}
\label{eq:cv}
CV(RMSE) = \frac{\sqrt{(1/n)\sum^n_{i=1} (y_i - \hat{y_i})^2}}{\bar{y}} * 100
\end{equation}

\begin{equation}
\label{eq:nmbe}
NMBE = \frac{(1/n)\sum^n_{i=1} (y_i - \hat{y_i})}{\bar{y}} * 100
\end{equation}

The best performing model is identified as the model that results in the lowest CV(RMSE), as this is the metric used to calculate the uncertainty introduced. The NMBE is used to support the model evaluation, however, it is not used in model selection due to the possibility of positive bias cancelling out negative bias and the objective of minimising modelling uncertainty. 

%\begin{equation}
%\label{eq:dist}
%Distance = \sqrt{CV(RMSE)^2 + NMBE^2}
%\end{equation} 

\subsection{Step 6 - Savings quantification}
\label{quant}
The previous steps are all completed prior to the implementation of the ECM. This ensures that any shortcomings in the approach are identified at an appropriate time that remedial actions can be taken. For example, if the feature selection algorithm showed that the variables were not strongly correlated to the dependent variable, then additional metering would be required to gather the necessary data in the baseline period. With these steps completed, the implementation period is used to implement the ECM and perform any commissioning works that are necessary. The final energy savings can then be quantified so long as the necessary reporting period data is available. 

\subsubsection{Data gathering}
Data for the final subset of model features must be gathered to enable calculation of the adjusted baseline. The information for data sources and associated data points logged in Section \ref{gathering} is used to guide this process. The data for both the model features and static factors must be gathered for the entirety of the reporting period. 

\subsubsection{Preprocessing}
The data gathered must be preprocessed to ensure integrity and accuracy is maintained. This preprocessing involves transforming the dataset into a format suited to the baseline period model. The contextualised feature names, measurement frequency and cleaning are all covered in this stage. As per ASHRAE Guideline 14, independent variables must not be more than 110\% of the maximum and no less than 90\% of the minimum values of the corresponding model training data. If the data does not conform to this requirement, an advisory note must be attached to the savings to state that the data is beyond the range of applicability of the model \cite{ASHRAE2014}.

As the model is trained on a standardised dataset, all input data must be standardised to the same scaling parameters. Therefore, the reporting period data must be scaled to the same mean and standard deviation as the training data set. This standardisation is required to maximise the performance of the modelling algorithms. 
% Note - IPMVP Core Concepts pg. 15 - "If any of the energy data are missing from the reporting period, a reporting period mathematical model can be created to fill in missing data. However, the reported savings for the missing period should identify these savings as missing data. 
% Too complicated. Keep it more straightforward.

\subsubsection{Model application}
The optimal model identified in Section \ref{eval} is applied to the prepared reporting period data to calculate the adjusted baseline. The scaling applied to standardise the reporting period data must now be reversed to give a context to the adjusted baseline. This can then be directly compared to the measured data for the same period of analysis. 

\subsubsection{Non-routine adjustments}
Non-routine adjustments are implemented on a project-by-project basis in circumstances with changes in static factors. These are often referred to as baseline adjustments (BLA). Each BLA is a custom engineering calculation for the given problem. They must be agreed between all project stakeholders. 

\subsubsection{Uncertainty}
The energy savings estimated must have an associated level of uncertainty and confidence. It is important to note that differing approaches for the calculation of uncertainty are proposed by the IPMVP and ASHRAE Guideline 14. In the IPMVP, acceptable uncertainty requires the savings to be larger than twice the standard error of the baseline value \cite{EfficiencyValuationOrganization2014}. ASHRAE Guideline 14 states that uncertainty must be less than 50\% of the annual reported savings, at a confidence level of 68\% \cite{ASHRAE2014}. The approach deemed suitable by the IPMVP is employed in this methodology. Thus, Equation \ref{eq:ipmvp} is applied, where $t$ is the t-statistic for a given level of confidence and degrees of freedom and $\text{SE}$ is the standard error of the estimate. 

\begin{equation}
	\label{eq:ipmvp}
	U = t * SE
\end{equation}

In the IPMVP, standard error is calculated using all of the baseline data. As discussed previously in Section \ref{partition}, this technique of using all baseline data to train and test the model is prone to over-fitting the model to that specific data set. The performance metrics are in turn calculated based on the models ability to fit the baseline data and subsequently, uncertainty is calculated. This can result in low levels of model error on the baseline data, but unreliable measures of uncertainty in the reporting period. The introduction of a random data split overcomes these issues by applying the model to an unseen testing data set.  

The ASHRAE approach to calculating uncertainty is not employed as it is too susceptible to the size of the baseline data set. It is found in a similar fashion to the IPMVP approach using the CV(RMSE) in Equation \ref{eq:ashrae}, which assumes that there is zero error introduced by the metering equipment; thus only the uncertainty introduced by model is calculated. The quantity of data available, required reporting period length and CV(RMSE) are the only variables influencing the quantity of uncertainty introduced in this phase. It is clear in this equation that as the quantity of training data available in the baseline period increases, the uncertainty reduces. The same is true for the length of the reporting period. To meet the uncertainty requirements already discussed, the required model accuracy reduces as the measurement frequency increases and the importance of an accurate baseline model is diminished. This is seen as a flawed approach that does not exude confidence in results. 

\begin{equation}
	\label{eq:ashrae}
	U = t * \frac{1.26 * CV(RMSE)}{F} * \sqrt{\frac{n + 2}{n * m}}
\end{equation}

\begin{figure}[h!]
	\centering
	\includegraphics[width=0.48\textwidth]{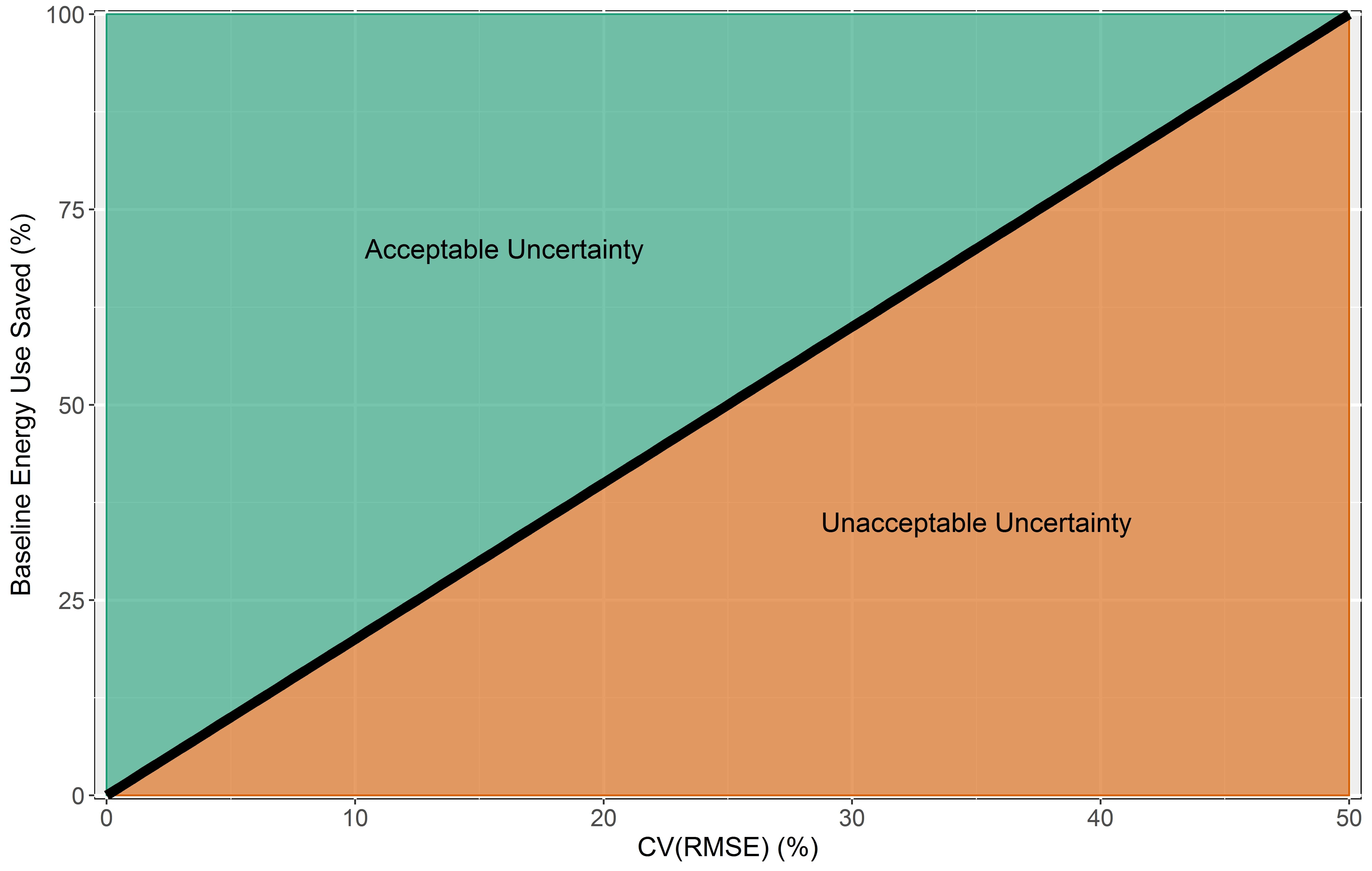}
	\caption{Model prediction performance requirements under varying fractional savings.}
	\label{lookup}
\end{figure}

The model performance required to ensure acceptable uncertainty changes relative to the quantity of savings resulting from an ECM. This relationship, based on the IPMVP approach, is illustrated in Figure \ref{lookup} and can be used as a reference chart for establishing the required performance levels. 

\section{Case study: Application of methodology and results}
\label{casestudy}
\begin{figure*}[]
	\centering
	\includegraphics[width=0.60\textwidth, keepaspectratio]{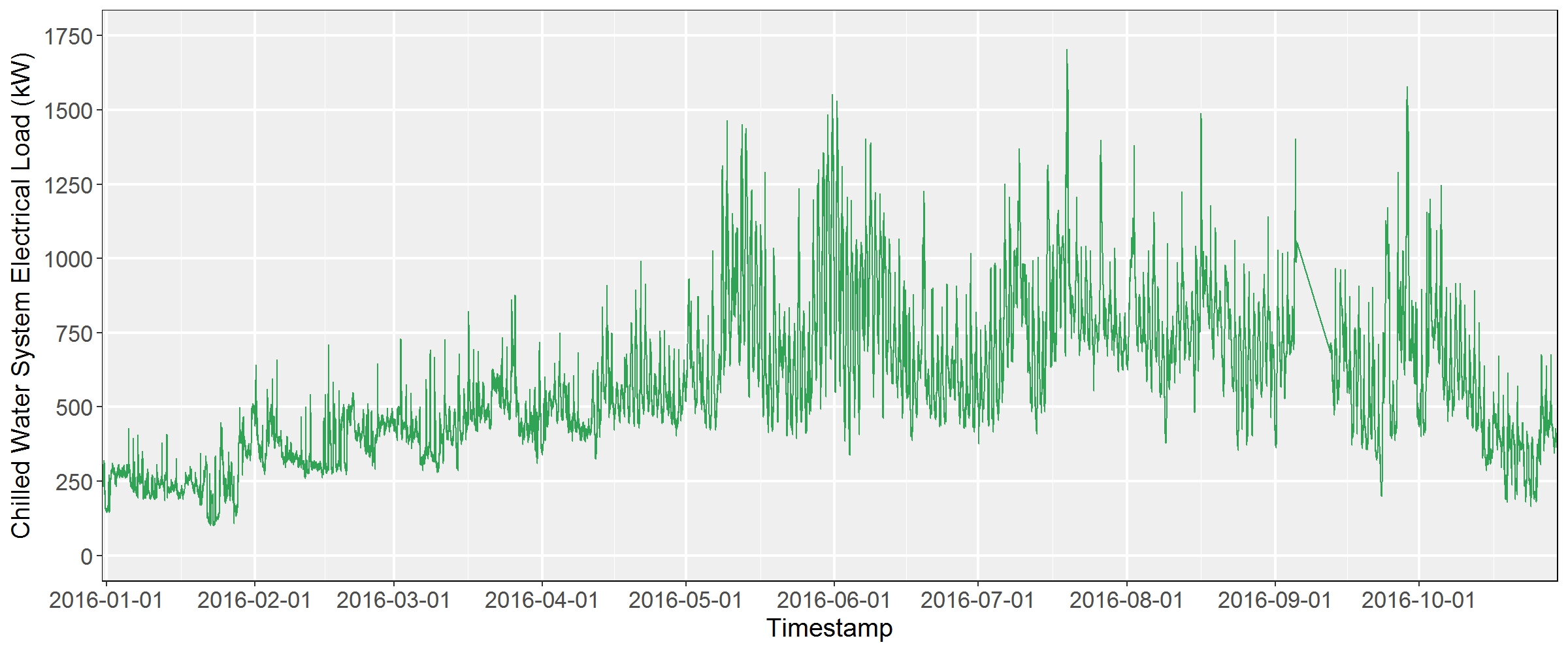}
	\caption{Electrical load of chilled water system in baseline period (pre-ECM).}
	\label{baseline}
\end{figure*}

The proposed methodology was applied to quantify the savings resulting from an ECM carried out on the chilled water system in a large biomedical manufacturing facility in Limerick, Ireland. The facility operates a continuous production process on a 24/7 basis. The generation and distribution of chilled water consumes approximately 7-8\% of the total site energy consumption annually. The chilled water generated by a series of electrically powered chillers is used to satisfy the space cooling loads across the facility. This is delivered using an array of air handling units. The ECM consisted of the reduction of chilled water consumption at an end-use level across the facility. This was achieved by identifying and optimising any air handling units that were operating outside of design specifications. Figure \ref{baseline} illustrates the energy consumption of the chilled water system prior to the implementation of the ECM, i.e. the baseline energy consumption.
  
The proposed approach was deemed suitable due to a number of project specific constraints. The whole-building approach outlined by the IPMVP was not suitable as the savings were estimated to be considerably less than 10\% of the total sites consumption. In addition to this, there was insufficient metering infrastructure to allow the successful application of retrofit isolation Options A or B. In residential and commercial buildings, it is often appropriate to utilise outside air temperature as the independent variable for modelling chilled water consumption, however, in this case the energy system has added complexity due to the production process in operation. The relationship between chilled water system electricity consumption and cooling degrees days was found to be weak with a coefficient of determination, $R^2$, of 0.36. In addition to this, the lack of availability of granular production data restricted analysis into the relationship between production output and chilled water consumption. To implement a conventional approach, additional metering would need to be installed, which in turn would delay the project implementation as baseline data would have to be gathered. 

Hence, the proposed methodology offers a novel alternative to the traditional approaches of M\&V. The methodology can be used to establish the relationships between total electrical consumption of the chilled water system and a variety of other metered quantities on site. Any significant relationships with independent variables can then be used to model the energy consumption in the baseline period, with a view to predicting the adjusted baseline following the implementation of the ECM. 

\subsection{Step 1 - Definition of project parameters}
Prior to the commencement of any ECM implementation works, the following project scope and parameters were defined:

\begin{itemize}
	\itemsep0em 
	\item \textbf{ECM}: Optimisation of air handling units to minimise the consumption of chilled water. This results in meeting the space cooling load with an increased efficiency.
	\item \textbf{Boundary}: The ECM will result in savings being achieved in the chilled water system electricity consumption. All other secondary benefits are outside the scope of this analysis.
	\item \textbf{Relevance to total facility consumption}: Chilled water system accounts for approximately 7-8\% of site electricity consumption.
	\item \textbf{Baseline period}: 1st January 2016 to 29th October 2016.
	\item \textbf{Implementation period}: 30th October 2016 to 15th February 2017.
	\item \textbf{Reporting period}: 16th February 2017 to 25th September 2017.
	\item \textbf{Relevant personnel}: Facilities engineering team, M\&V practitioner.
	\item \textbf{Data sources}: Building management system (BMS), energy management software (EMS).
	\item \textbf{Static factors}: Number of production lines, size of facility, production process, shift scheduling, building fabric, space heating and cooling set-points, air change rates. 
\end{itemize}
%Include a schematic of the chilled water system? Most likely not allowed for client confidentiality. 

\subsection{Step 2 - Data gathering}
The existing metering infrastructure was utilised to develop a model for the chilled water system electricity consumption. Both the BMS and EMS store valuable data gathered by electrical, mechanical and climatic meters located across the facility. The characteristics of each data source are documented in Table \ref{datasources}. This will be referred to in the reporting period to replicate the data gathering process.  

\begin{table}[h!]
	\small
	\def\arraystretch{1.5}
	\begin{tabular}{>{\raggedright}p{0.17\textwidth}>{\raggedright}p{0.13\textwidth}p{0.12\textwidth}}
		\hline
		Characteristic & BMS & EMS\\
		\hline
		Type of data & Mechanical, electrical \& climatic & Electrical\\
		Measurement frequency & Varies & 15-min\\
		Storage & On-site server & Remote server\\
		Access & Local network access & Cloud access\\
		\hline
	\end{tabular}
	\caption{Characteristics of data sources.}
	\label{datasources}
\end{table}

\begin{figure}[h!]
	\centering
	\includegraphics[width=0.5\textwidth]{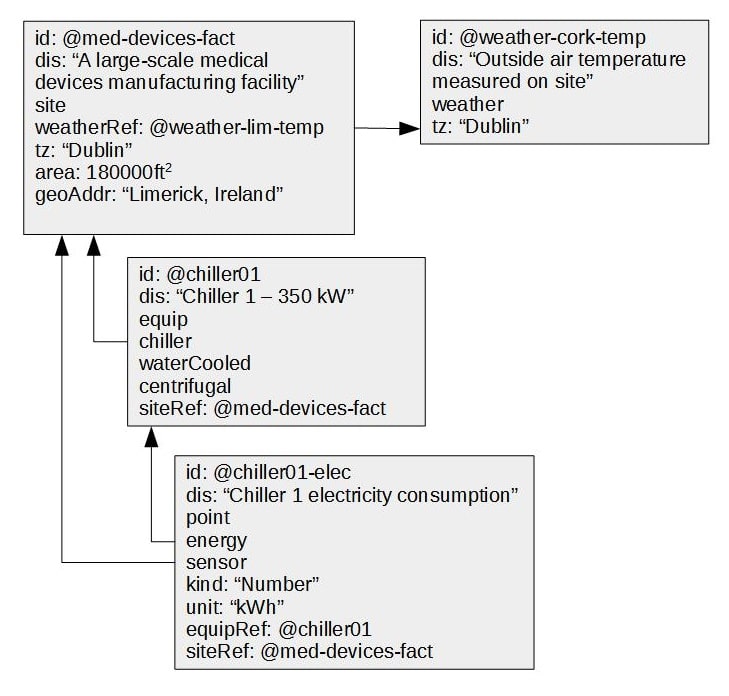}
	\caption{Example of the application of the haystack naming convention to the data.}
	\label{haystackapp}
\end{figure}

As discussed in Section \ref{gathering}, poor semantic modelling of energy data in industrial facilities is common. The Haystack naming convention was applied to the dataset to ensure each data point has a context with regard to the site. This increases the ease with which the data-driven model is then applied to data gathered in the reporting period. Figure \ref{haystackapp} gives an example of the application of the Haystack naming convention to a sample data point. The two data sources were then collated together into a single dataset of contextualised data points. 

\subsection{Step 3 - Feature selection}
The dataset available for analysis contained 505 variables, each corresponding to a unique physical meter on-site. When the dependent variable, the chilled water system electricity consumption, is removed from this set, there are 504 variables that can be input to the model development process. The use of all of these variables to construct a model of the dependent variable would not be sensible given the computing resources typically available to practitioners. All variables used to model the consumption must add significant value to the model to be considered statistically significant. 

Application of the feature selection algorithms outlined in Section \ref{featsel} were applied to the dataset. This resulted in the identification of 15 variables that added value to the multiple regression model explaining the dependent variable. Collinearity tests were conducted to ensure all features are independent variables and multicollinearity did not exist. 

\subsection{Step 4 - Availability assessment and cleaning}
\begin{figure*}[h!]
	\centering
	\includegraphics[width=0.95\textwidth, keepaspectratio]{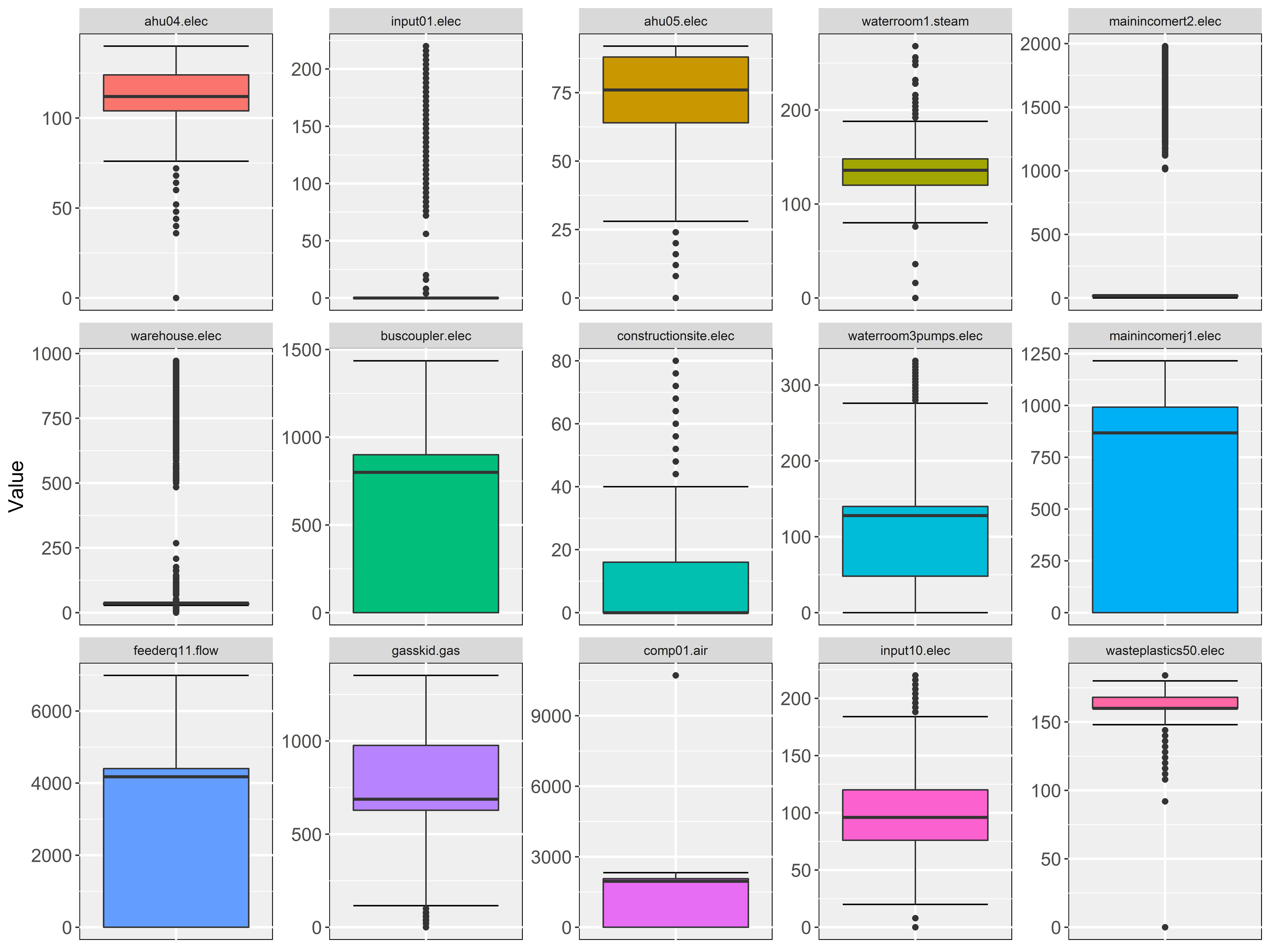}
	\caption{Box and whisker plots generated to evaluate each proposed model feature.}
	\label{boxplots}
\end{figure*}

Box and whisker plots were used to visualise the statistical measures detailed in Section \ref{avail}. These plots are included in Figure \ref{boxplots}. Any points plotted outside the whiskers are considered outliers. Outliers are classified as being greater or less than 1.5 times the inter quartile range. 

% Would be far better as a plot
%\begin{table*}[h!]
%	\small
%	\def\arraystretch{1.5}
%	\begin{tabular}{>{\raggedright}p{0.17\textwidth}>{\raggedright}p{0.07\textwidth}p{0.07\textwidth}>{\raggedright}p{0.1\textwidth}>{\raggedright}p{0.09\textwidth}>{\raggedright}p{0.07\textwidth}>{\raggedright}p{0.07\textwidth}p{0.07\textwidth}p{0.07\textwidth}}
%		\hline
%		Feature & Mean & Median & No. of Unique Values & No. of Missing Value & $1^{st}$ Quartile & $2^{nd}$ Quartile & Minimum & Maximum \\
%		\hline
%		ahu04-elec & 111.5 & 112 & 27& 0 & 104 & 124 & 0 & 140 \\
%		input01-elec & 19.05& 0 & 44& 0 & 0 & 0 & 0 & 220 \\
%		ahu05-elec & 74.15& 76 & 23& 0 & 64 & 88 &0 &92\\
%		waterroom1-steam & 134.5 & 136 & 45 & 0 & 120& 148 & 0& 268\\
%		mainincomert2-elec & 218.3 & 16 & 209 & 0 & 12& 20&  0& 1980\\
%		warehouse-elec & 107.1& 36 & 151 & 0 &32 &36 &  0 &972 \\
%		buscoupler-elec & 531 & 800 & 257& 0 & 0& 900&  0& 1436 \\
%		constructionsite-elec & 7.35& 0 & 21& 0 & 0& 16&  0& 80 \\
%		waterroom3pumps-elec & 111.3 & 128 & 83& 0& 48 & 140 &  0& 332\\
%		mainincomerj1-elec & 658 & 868 & 179& 0 & 0 & 992 &  0& 1216\\
%		feederq11-flow & 2987 & 4176 & 742& 0 & 0 & 4404 &  0 & 6988\\
%		gasskid-gas & 765 & 688 & 278 & 0 & 628 & 976 &  0 & 1352\\
%		comp01-air & 1161 & 1960 & 136& 0 & 0 & 2068 &  0 & 10716\\
%		input10-elec & 100.2 & 96 & 49& 0 & 76 & 120 &  0 & 220\\
%		wasteplastics50-elec & 162.6 & 160 & 22 & 0 & 160 & 168 &  0 & 184\\
%		\hline
%	\end{tabular}
%	\caption{Summary statistics of feature output from feature selection algorithm.}
%	\label{app_sumstats}
%\end{table*}

The results of the availability assessment are used to identify which features require cleaning. To comply with the IPMVP practices, this cleaning simply consists of omitting features identified as unclean from the analysis. No back-filling of unclean data is permissible. Any feature that had more than 5\% of data identified as outliers were omitted from the feature set. 

As a result of the data cleaning, 10 features were identified as being suitable for baseline energy model training. These features, or independent variables, selected to model the chilled water system electricity consumption are detailed in Table \ref{features}. For confidentiality reasons, the fully contextualised feature names have not been included. 

\begin{table}[h!]
	\small
	\def\arraystretch{1.5}
	\begin{tabular}{>{\raggedright}p{0.14\textwidth}p{0.29\textwidth}}
		\hline
		Feature & Description \\
		\hline
		ahu04-elec & Electricity consumption of air handling unit no. 4.\\
		ahu05-elec & Electricity consumption of air handling unit no. 5.\\
		constructionsite-elec & Electricity consumption of construction site.\\
		waterroom3pumps-elec & Electricity consumption of pumps in water room no. 3.\\
		mainincomer-elec & Electricity consumption recorded on site incomer from grid.\\
		feederq11-elec & Production related electricity consumption.\\
		gasskid-gas & Gas consumption of gas skid.\\
		comp01-air & Compressed air produced by compressor no. 1\\
		input10-elec & Production related electricity consumption.\\
		wasteplastics50-elec & Electricity consumption related to processing of production waste.\\
		\hline
	\end{tabular}
	\caption{Feature set output from application of feature selection algorithm.}
	\label{features}
\end{table}

The $R^2_{adjusted}$ value for these 10 features was 0.663. Although no strict limit for acceptable $R^2_{adjusted}$ exists, this value is on the lower end of acceptable values. This value does however increase as measurement frequency decreases indicating that the relationship is stronger as data granularity increases. This could possibly be due to delays between variable responses that do not cause an issue with lower measurement frequency.  

\subsection{Step 5 - Baseline energy modelling}
The initial dataset gathered has been minimised to include the chosen independent variables (features) and the dependent variable; this is known as the feature set. The feature set was gathered using a 15-minute measurement frequency. The data was aggregated to create four independent feature sets with 15-minutes, hourly, daily and weekly measurement frequencies. This enables an exhaustive approach to modelling, which results in the identification of the optimal data granularity. 

Each feature set was then split into training and testing data using an 80:20 ratio. This allows models to be evaluated on unseen data, hence, improving reliability of results. The training data was then standardised to ensure the best possible model fit is achieved. An OLS, k-NN, ANN and SVM regression model was trained for each measurement frequency. A grid search approach was used to train the hyper-parameters of each model as discussed in Section \ref{modtraining}. This exhaustive approach led to 16 models being constructed. 

\begin{figure}[]
\centering
\includegraphics[width=0.49\textwidth]{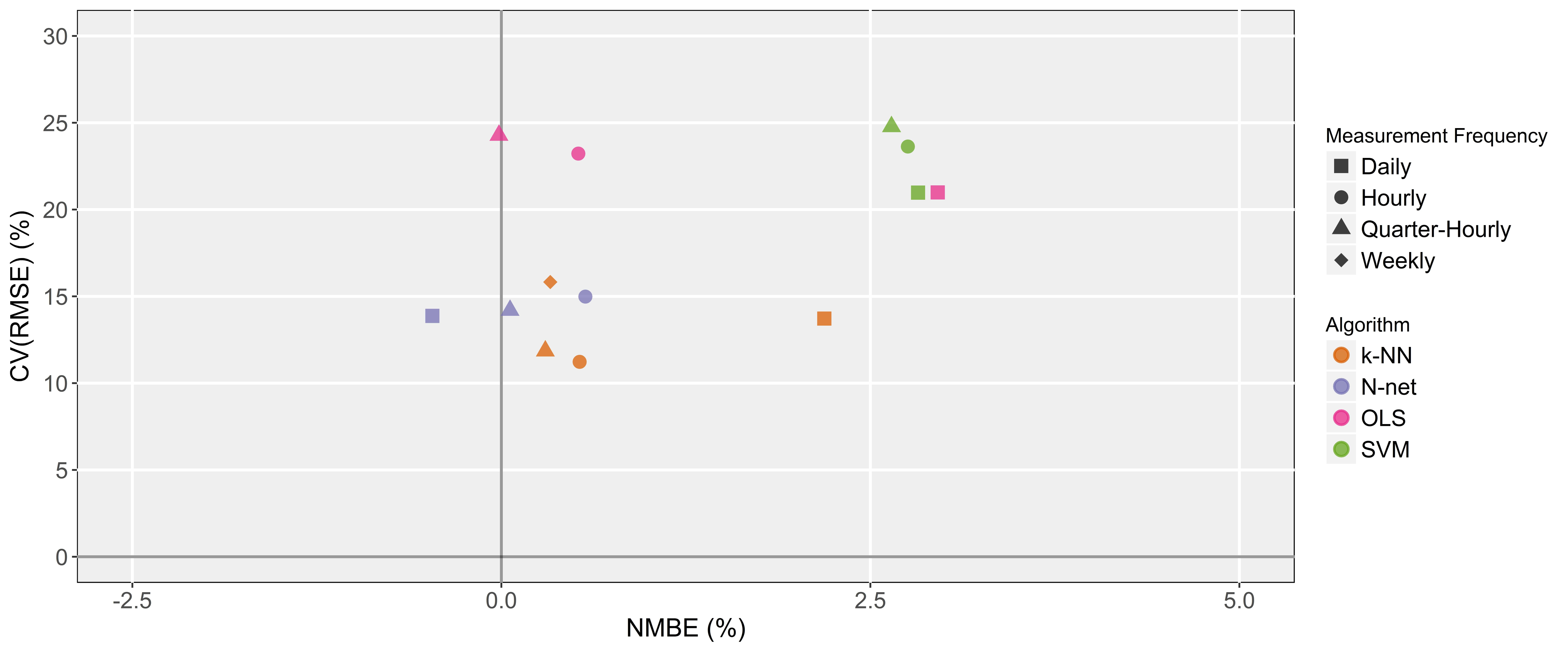}
\caption{Performance of all models evaluated on testing data set.}
\label{testingperf}
\end{figure}

In contrast to practices employed by the IPMVP and ASHRAE Guideline 14, unseen data was used to evaluate the performance of each model in the baseline period. This helps prevent over-fitting the model to the training data set, thus increasing its applicability. The standard error, or RMSE, is used to calculate uncertainty in the final savings, hence, it is important that this is reliably quantified. The testing data set was normalised using the same scaling factors applied to the training data set. The performance was then evaluated by applying each model constructed to the testing data set and comparing estimated values of the dependent variable to measured values. This would not be possible without the data partition already carried out. Figure \ref{testingperf} illustrates the performance of each model in predicting the chilled water system electricity consumption during the baseline period. 

\label{appmodel}

The optimal model was selected based on CV(RMSE), as this metric directly impacts on energy savings uncertainty in the reporting period. A k-NN model trained using data with an hourly measurement frequency was the best performing model with a CV(RMSE) of 11.23\%. A triangular kernel, 5 being the maximum number of neighbours and a distance equal to 1 were the associated model hyper-parameters. 

\subsection{Step 6 - Savings quantification}
Following the complete implementation of the ECM, the reporting period began. The information relating to data sources and characteristics documented in Sections \ref{define} was used to gather the data necessary to quantify savings in the reporting period. It is at this point that the value of having contextualised the data comes to fruition as the data can be gathered more easily with clear semantic modelling. As with the baseline period, the raw data is gathered from both data sources and stored in a single data set. 

Preprocessing consists of checking the quality of the data and aggregating it into the measurement frequency corresponding to the optimal baseline model. Quality checking is performed by assessing if the independent variables data gathered in the reporting period conforms to the range requirements of ASHRAE Guideline 14 (i.e. must not be more than 110\% and no less than 90\% of the corresponding baseline data). This was not an issue for this particular case study. The data was measured with a 15-minute measurement frequency and was subsequently aggregated to have an hourly measurement frequency to conform to the baseline energy model input requirements. The model was then applied to the processed data set to calculate the adjusted baseline energy consumption.

In most cases, the adjusted baseline can be directly compared to measured quantities of the dependent variable for quantification of savings. However, this is not the case in this application as there was a change in static factors during the M\&V period of analysis. The construction of a production area occurred during the implementation period and was live at the beginning of the reporting period. This area houses an additional production line and has the capacity for the operation of additional production lines in the future. This increased the cooling load of the facility which requires the savings calculated be adjusted. While the chilled water system electrical consumption increased, the cooling load also increased. Hence, had the ECM not been implemented, the load would have been satisfied in a less efficient manner with even higher chilled water system electrical consumption. This non-routine adjustment was made based on the floor area of the site increasing by 20\%. This was deemed acceptable as the chilled water system services the space cooling requirements. 

The energy savings in the reporting period are the difference between the measured consumption and the adjusted baseline, following the application of the non-routine adjustment. The savings were calculated to be 604,527 kWh. A mere calculation of savings without an associated level of uncertainty and confidence is of little use in ensuring reliability and completeness in M\&V. The range of savings must be calculated for a given confidence interval to gain a true insight into project performance. This range is dependent on the model performance in the baseline period. The more accurate the baseline energy model, the smaller the range of savings. The IPMVP approach to uncertainty calculation detailed in Equation \ref{eq:ipmvp} was employed to calculate the range of savings in this project to be;

$$\text{Range of Savings} =  604,527 \pm \text{(t x S.E.)}$$
$$ = 256,485 \text{ to } 952,568 \text{ kWh @ 68\% Confidence}$$ 

\section{Discussion}
\label{results}
\subsection{Energy savings in reporting period}
\begin{figure}[!h]
	\centering
	\includegraphics[width=0.47\textwidth]{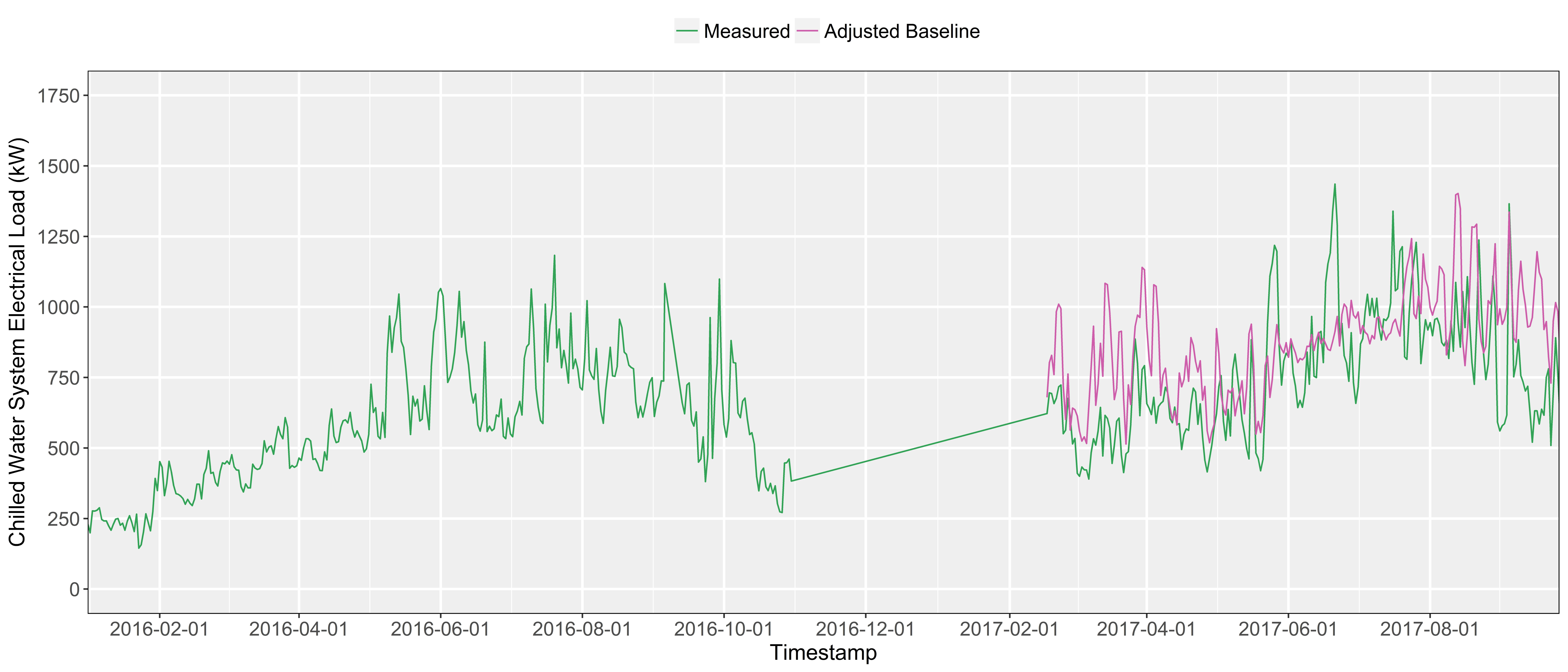}
	\caption{Measured consumption and adjusted baseline for entire period of analysis.}
	\label{alldata}
\end{figure}

The savings were calculated as being the difference between measured consumption in the reporting period and the adjusted baseline. This is illustrated in Figure \ref{alldata}. The regression model is capable of forecasting well relative to the measured data for large parts of the reporting period, although there are periods in the summer months of 2017 in which the model actually predicts that more energy is being consumed than would have been pre-ECM. 

\begin{table}[h!]
	\small
	\def\arraystretch{1.5}
	\begin{tabular}{>{\raggedright}p{0.09\textwidth}>{\raggedright}p{0.07\textwidth}>{\raggedright}p{0.065\textwidth}>{\raggedright}p{0.065\textwidth}p{0.065\textwidth}}
		\hline
		Measurement \newline Frequency & Algorithm & Minimum (kWh) & Median (kWh) & Maximum (kWh)\\
		\hline
		\multirow{4}{*}{15-min} & OLS & -113,236 & 628,938 & 1,371,112 \\
			& k-NN & 126,907 & 489,202 & 851,497 \\
			& ANN & 747,769 & 1,181,674 & 1,615,579 \\
			& SVM & -335,004 & 422,329 & 1,179,662 \\
		\hline
		\multirow{4}{*}{Hourly} & OLS & -60,369 & 649,542 & 1,359,454 \\
			& k-NN & 261,399 & 604,527 & 947,655 \\
			& ANN & 687,936 & 1,145,835 & 1,603,735 \\
			& SVM & -273,881 & 448,371 & 1,170,623 \\
		\hline
		\multirow{4}{*}{Daily} & OLS & -7,055 & 654,115 & 1,315,285 \\
			& k-NN &  123,428 & 555,656 & 987,884  \\
			& ANN & 346,580 & 783,606 & 1,220,632  \\
			& SVM & -73,504 & 587,279 & 1,248,061 \\
		\hline
		\multirow{4}{*}{Weekly} & OLS & 285,067 & 814,873 & 1,344,678 \\
			& k-NN & -82,744 & 402,815 & 888,374 \\
			& ANN & -278,097 & 644,592 & 1,567,281 \\
			& SVM & 786,834 & 1,355,583 & 1,924,333 \\
		\hline
	\end{tabular}
	\caption{Savings in kWh for all models developed under varying measurement frequency with a confidence interval of 68\%.}
	\label{tab:range}
\end{table}

\subsection{Range of savings}
\begin{figure*}[!h]
	\centering
	\includegraphics[width=0.95\textwidth, keepaspectratio]{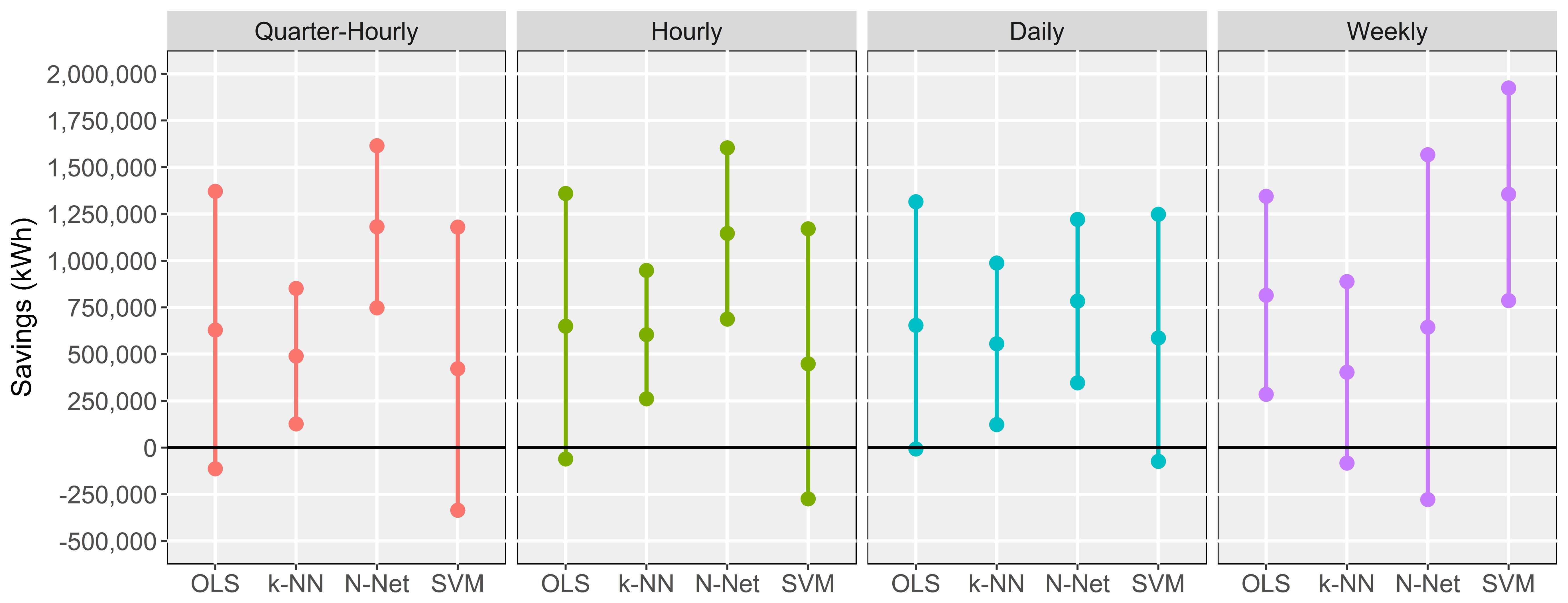}
	\caption{Range of savings for all models developed under varying measurement frequency with a confidence interval of 68\%.}
	\label{savingsrange}
\end{figure*}

The impact of modelling error on the range of savings can be seen in Figure \ref{savingsrange} with detailed results provided in Table \ref{tab:range}. The results include the savings estimated by all models developed and the associated range of savings estimated with 68\% confidence. The uncertainty in a project can be directly seen in the range associated with the quantified savings. The mean savings across all models is 710,558 kWh with a coefficient of variation (CV) of 39.9\%. Although the savings quantified by each of the 20 models are dependent on the measurement frequency, algorithm and individual hyper-parameters, trends are evident over the set of models. The ANN regression models predict the highest savings on average in each frequency subset with an average savings of 938,927 kWh and a CV of 28.3\% . The OLS models are the most consistent across all measurement frequencies with an average savings figure of 686,866 kWh and a CV of 12\%, however, these savings also had the largest associated ranges. In terms of average savings, the SVM models were the most erratic with a CV of 62.6\% across results. 

This analysis shows the sensitivity of baseline energy models to algorithms, measurement frequencies and hyper-parameters. The range of savings across all 20 models is particularly distressing and this emphasises the need to maximum accuracy in baseline energy models to ensure the range of savings are small; thus, maximising confidence in savings. 

\subsection{Acceptable uncertainty}
As stated in the IPMVP, uncertainty is deemed acceptable when the savings are larger than twice the standard error of the baseline energy model. For comparative purposes, each model developed was assessed to check if the uncertainty levels could be deemed acceptable. Table \ref{acceptable} contains the results of this analysis. It was found that only 3 of the 20 models developed meet the criteria defined by the IPMVP. Critically, the optimal model identified and applied for the final calculation of savings in the case study (k-NN with hourly measurement frequency) does not meet the criteria for acceptable uncertainty. The problem with this check is that the models that tend to predict a higher quantity of savings are favoured, without necessarily being the most accurate models. As discussed, the ANN models predicted the highest savings on average across all measurement frequencies, while the k-NN models had to lowest CV(RMSE) for every measurement frequency. This results in the k-NN models not being deemed acceptable due to their more conservative estimation of savings. Hence, this is somewhat of a flawed process for the final evaluation of model performance. 

\begin{table}[h!]
	\small
	\def\arraystretch{1.5}
	\begin{tabular}{>{\raggedright}p{0.09\textwidth}>{\raggedright}p{0.065\textwidth}>{\raggedright}p{0.07\textwidth}>{\raggedright}p{0.07\textwidth}p{0.07\textwidth}}
		\hline
		Measurement\newline Frequency & Algorithm & Savings\newline (kWh) & Standard Error\newline (kWh) & Acceptable\\
		\hline
		15-min & OLS & 628,938 & 746,293 & No\\
		15-min & k-NN & 489,202 & 364,306 & No\\
		15-min & ANN & 1,181,674 & 436,313 & Yes\\
		15-min & SVM & 422,329  & 761,536 & No\\
		Hourly & OLS & 649,542 & 713,806 & No\\
		Hourly & k-NN & 604,527 & 345,010 & No\\
		Hourly & ANN & 1,145,835 & 460,412 & Yes\\
		Hourly & SVM & 448,371  & 726,214 & No\\
		Daily & OLS & 654,115& 663,477 & No\\
		Daily & k-NN & 555,656 & 433,737 & No\\
		Daily & ANN & 783,606 & 438,551 & No\\
		Daily & SVM & 587,279 & 663,088 & No\\
		Weekly & OLS & 814,873 & 525,191 & No\\
		Weekly & k-NN & 402,815 & 481,329 & No\\
		Weekly & ANN & 644,592 & 914,651 & No\\
		Weekly & SVM & 1,355,583 & 563,795 & Yes\\
		\hline
	\end{tabular}
	\caption{Acceptable levels of uncertainty for each model developed.}
	\label{acceptable}
\end{table}

\subsection{Measurement uncertainty}
The uncertainty in the savings quantified is due to the error introduced by the baseline energy modelling. Measurement error was omitted from this analysis to analyse the performance of the regression model in isolation. It is critical that the range of savings reported is minimised to maximise confidence in the M\&V process. It is important to note that the measurement error is likely to increase as more independent variables are required to model the baseline energy consumption. In particular, when these measurements are recorded using metering infrastructure on-site, each will have an associated uncertainty. This is in contrast to the more simplistic modelling solutions that rely on weather and occupancy data. 

%\subsection{Non-routine adjustment}
%The increasing chilled water system electrical load over the course of the project required the application of a non-routine adjustment, thus increasing the complexity of the task. This requires knowledge of both the energy and production systems causing it to be difficult to formulate explicit guidance on. It is important to rely on statistical measures to ensure any adjustments made are valid and can be relied upon. 

\section{Conclusions}
\label{concs}
%Methodology
The research presented in this article offers a novel approach to utilise machine learning techniques for energy savings verification. A focus is placed on minimising uncertainty introduced by the energy model in the quantified savings. A definitive methodology was developed to provide explicit guidance on the application of machine learning for the purposes of maximising the accuracy with which M\&V can be carried out. M\&V practitioners do not need to have knowledge of the inner working of the modelling algorithms, as the step-by-step approach to the problem includes performance checks that must be met to ensure accuracy. A whole-facility approach is adopted to widen the scope of analysis, while isolating the energy system in which the ECM is implemented. This blended approach defines a novel boundary of analysis and makes use of all data recorded across the facility, without requiring the energy savings to be large relative to the site load (IPMVP Option C constraint). The methodology has been designed to be robust enough to be compatible across the spectrum of M\&V projects. 

%Case Study
The proposed methodology is of particular benefit in circumstances with limited metering infrastructure directly related to the energy system under analysis. Evidence of which can be seen in the case study where M\&V would not have been possible without the installation of additional metering equipment and a data gathering period that would delay the ECM implementation. The methodology was directly applied using real-world data for a large biomedical manufacturing facility. A total of 20 models were developed in the baseline period using an exhaustive approach. The optimal model was identified as a k-NN regression model trained with data measured hourly. The CV(RMSE) of this model was 11.23\%. The reporting period was 222 days in duration with estimated energy savings over this period being 604,527 kWh. Critically, following the quantification of the associated uncertainty, these savings were found to range from 256,485 to 952,568 kWh at 68\% confidence. The range of savings for all 20 models constructed were investigated to show the impact model performance has on final savings. This highlighted that the procedure used to assess acceptable uncertainty favours models that estimate higher savings in the reporting period, rather than those that perform better in the independent cross validation testing.

%Case Study Problems
It is important to note that the conditions in the case study are representative of the issues that exist across the industrial sector. These include a lack of sufficient metering, poor data quality and changing static factors. The application of the methodology demonstrates its ability to overcome these issues and quantify energy savings with an acceptable level of uncertainty. The level of uncertainty achievable is dependent on individual project characteristics. Specifically, the relationship between the dependent variable and available independent variables is the limiting factor in minimising uncertainty. The proposed methodology seeks to realise the achievable uncertainty by utilising the available resources. In the case study, the industrial building did not represent ideal conditions and the results illustrate this. The relationship between the chilled water electrical load and the 10 independent variables had an adjusted $R^2$ value of 0.663. This was for a 15-minute measurement frequency which represented the poorest value of all the frequencies analysed. As this is on the lower end of statistically significant values, the achievable model performance is limited by this relationship. Evidence of this can be seen in the large range of savings in the final results. The problematic nature of the case study enabled a robust assessment of the methodology to take place. 

%Resources reduced
The approach taken by the methodology reduces the need to install additional metering infrastructure. This has the potential to greatly reduce the resources required to complete accurate M\&V in any given project. Machine learning techniques are able to extract relevant knowledge from the available data set and utilise it to develop the baseline energy model. 

\section{Future work}
\label{future}
Future work will focus on incorporating the modelling methodology into an M\&V framework for real-time, automated savings verification. The objective is to develop a framework that can seamlessly handle the transition from real-time M\&V to long-term monitoring and targeting (M\&T). Persistence of savings has come under the spotlight with the implementation of the EU Energy Efficiency Directive and the tools on the market to solve this problem are very much residential and commercial buildings focused. This framework will offer an M\&V 2.0 software solution for industrial buildings that will enable a seamless transition to M\&T ensuring persist savings are realised. 
 
\section*{Acknowledgements}
The authors would like to acknowledge the funding received by the Science Foundation Ireland under Grant no. 12/RC/2302.
%The authors would like to acknowledge the funding received from the Science Foundation Ireland funded Marine and Renewable Energy Research Centre and the NTR Foundation (under Grant XX/123/XXXX) with whom this research was undertaken.

\section*{References}
\bibliographystyle{elsarticle-num}

\end{document}